\documentclass[conference]{IEEEtran}
\IEEEoverridecommandlockouts

\usepackage{cite}
\usepackage{amsmath,amssymb,amsfonts}
\usepackage{algorithmic}
\usepackage{graphicx}
\usepackage{textcomp}
\usepackage{xcolor}
\usepackage{xurl}
\usepackage{subcaption}
\usepackage{hyperref}
\usepackage{multirow}
\usepackage{fancyhdr}

\def\BibTeX{{\rm B\kern-.05em{\sc i\kern-.025em b}\kern-.08em
    T\kern-.1667em\lower.7ex\hbox{E}\kern-.125emX}}

\makeatletter
\def\headrule{
    \vskip-\headrulewidth
    \hrule height \headrulewidth
    \vskip-\headrulewidth
    \vskip 0.2cm  
    \hbox to \headwidth{%
        \hfil\textit{This paper was accepted at CBMI 2025} \hfil
    }
}
\makeatother

\begin{document}

\title{AgriPotential: A Novel Multi-Spectral and Multi-Temporal Remote Sensing Dataset for Agricultural Potentials
}

\author{\IEEEauthorblockN{Mohammad El Sakka}
\IEEEauthorblockA{\textit{IRIT UMR5505 CNRS} \\
\textit{University of Toulouse}\\
Toulouse, France \\
0009-0007-0531-7946}
\and
\IEEEauthorblockN{Caroline De Pourtales}
\IEEEauthorblockA{\textit{IRIT UMR5505 CNRS} \\
\textit{CNRS PNRIA}\\
Toulouse, France \\
caroline.de-pourtales@irit.fr}
\and
\IEEEauthorblockN{Lotfi Chaari}
\IEEEauthorblockA{\textit{IRIT UMR5505 CNRS} \\
\textit{Toulouse INP}\\
Toulouse, France \\
0000-0002-3590-0370}
\and
\IEEEauthorblockN{Josiane Mothe}
\IEEEauthorblockA{\textit{IRIT UMR5505 CNRS} \\
\textit{INSPE, UT2J, Univ. de Toulouse}\\
Toulouse, France \\
0000-0001-9273-2193}
}
\IEEEaftertitletext{\vspace{-2\baselineskip}}

\maketitle

\fancypagestyle{firstpage}{
  \fancyhf{}
  \fancyhead[L]{
    \parbox[c][2cm][c]{5cm}{
      \centering
      \includegraphics[height=2cm]{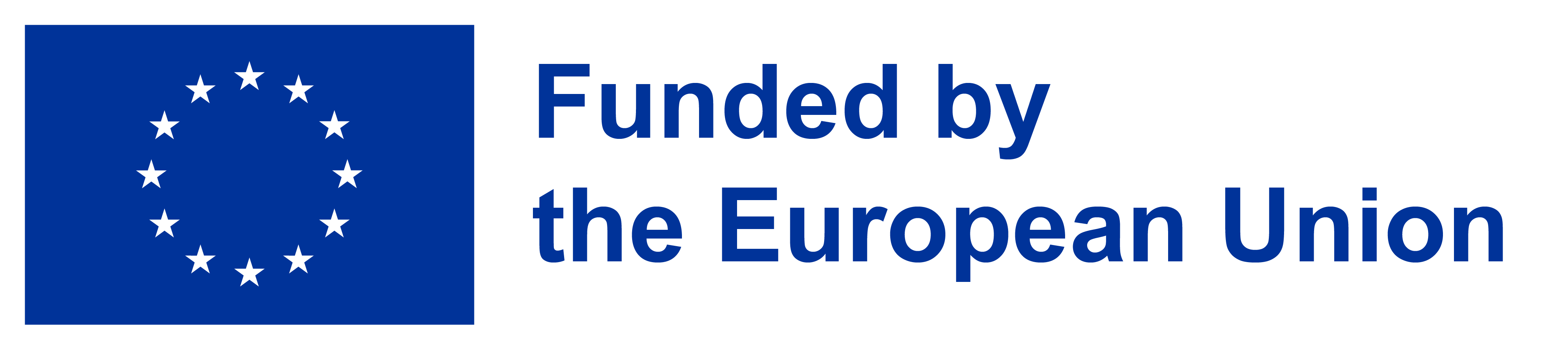}
    }
  }
  \fancyhead[R]{
    \parbox[c][2cm][c]{8cm}{
      \raggedright
      Views and opinions expressed are those of the authors only and do not necessarily reflect those of
      the European Union. Neither the European Union nor the granting authority can be held responsible for them.
    }
  }
 \setlength{\headheight}{2cm} 
}

\thispagestyle{firstpage}

\begin{abstract}
Remote sensing has emerged as a critical tool for large-scale Earth monitoring and land management. In this paper, we introduce AgriPotential, a novel benchmark dataset composed of Sentinel-2 satellite imagery captured over multiple months. The dataset provides pixel-level annotations of agricultural potentials for three major crop types - viticulture, market gardening, and field crops - across five ordinal classes. AgriPotential supports a broad range of machine learning tasks, including ordinal regression, multi-label classification, and spatio-temporal modeling. The data cover diverse areas in Southern France, offering rich spectral information. AgriPotential is the first public dataset designed specifically for agricultural potential prediction, aiming to improve data-driven approaches to sustainable land use planning. The dataset and the code are freely accessible at: \url{https://zenodo.org/records/15551829}
\end{abstract}


\section{Introduction}

Remote sensing is a powerful means of monitoring the Earth's surface. With more advanced satellites that incorporate high-resolution spectral, spatial, and temporal instruments~\cite{zhao2022overview, zhang2022progress}, remote sensing data have become increasingly valuable for land mapping~\cite{macarringue2022developments}, change detection~\cite{gu2024use}, and other types of predictions~\cite{zhang2022artificial}. To extend progress in this domain, we introduce AgriPotential, a novel benchmark dataset built from multi-spectral multi-temporal Sentinel-2 imagery. Specifically, the dataset is built for agricultural potential prediction, offering pixel-level annotations across five ordinal classes and three crop types. Unlike many existing remote sensing datasets that typically target a single machine learning task, AgriPotential is well-suited for a diverse set of learning tasks, including ordinal regression, single and multi-label classification, regression, and spatio-spectral-temporal modeling. Additionally, to the best of our knowledge, there are no publicly available datasets that focus on predicting agricultural potentials, making AgriPotential the first of its kind.

 \begin{table}
    \centering
    \caption{AgriPotential dataset summary.}
    \label{tab:specs}
    \begin{tabular}{l|c}
         Property & Value  \\
         \hline
         Dataset name & AgriPotential \\ 
         File format & HDF5 (.h5) \\ 
         File size & 28.4 GB \\ 
         Data volume & 145.7 GB \\ 
         Number of images & 8,890 \\
         Temporal length & 11 \\
         Spectral channels & 10 \\
         Spatial resolution & 5 m/px \\
         Dimensions & 128x128 (0.41 km$^2$) \\
         Annotation level & Pixel level \\
         Crop types & viticulture, market gardening, field crops \\
         Potential classes & Very low, low, average, high, very high \\
         License &  CC BY 4.0 \\
         Data link & \url{https://zenodo.org/records/15551829} \\
         \hline
    \end{tabular}
\end{table}

Agricultural potential - often referred to as agricultural productivity, soil capability, capacity, or crop suitability - is the ability of a specific area to support agricultural production~\cite{fao_-_food_and_agriculture_organization_of_the_united_nations_6_nodate}. Determining the potentials is fundamental to sustainable land management and agricultural planning, as it enables decision-makers to optimally use land by selecting the most suitable crops for a given area, which optimizes the use of natural resources, and improves overall productivity and sustainability~\cite{fao_-_food_and_agriculture_organization_of_the_united_nations_6_nodate}. Traditionally, the assessment of agricultural potentials is a manual and observation-based process. This involved field surveys, in-situ measurements, soil testing, historical patterns, and trial and error~\cite{fao_-_food_and_agriculture_organization_of_the_united_nations_6_nodate}. Recently, some studies have addressed this problem using more systematic data-driven methods. These studies often analyze soil chemicals, nutrient levels, pH and weather conditions based on in-situ measurements and leveraging IoT and cloud technologies. These methods  are labor-intensive and time-consuming, primarily carried out by dedicated experts, and often lack field validation~\cite{ed2023predictive, el2022assessment, mcbratney2019soil, alnaimy2022spatio, radovcaj2022gis, bhullar2023simultaneous}.

To address these challenges, satellite imagery is presented as an alternative source of information that enables large-scale and up-to-date assessment of agricultural potentials without the need for constant in-situ data collection. Among the various types of satellite sensors - such as thermal, radar, hyperspectral, and multispectral - we explore the possibility of using multispectral instruments due to their proven effectiveness and widespread availability~\cite{elmasry2019recent, ma2023multispectral, kumari2023sentinel}. Multispectral sensors capture the reflected radiation from the Earth's surface across multiple wavelength bands - typically up to 20 - within the electromagnetic spectrum. Many studies have demonstrated the effectiveness of combining machine learning approaches with multi-spectral satellite remote sensing in tasks related to Land Use and Land Cover applications~\cite{cong2022satmae, karra2021global, tikuye2023land}  and agriculture~\cite{el2024images, el2025review, segarra2020remote, singh2022deep}. Incorporating multi-temporal images enables the observation of temporal patterns and land dynamics, which are essential in tasks related to Earth Observation.

AgriPotential integrates multi-spectral and multi-temporal satellite imagery, spanning diverse agricultural regions in Southern France. It covers three major crop types: viticulture, market gardening, and field crops.

While AgriPotential is built around an agricultural topic, its structure and properties make it valuable for a broad range of remote sensing and machine learning research. A summary of the dataset's specifications is provided in Table~\ref{tab:specs}. The key contributions of this paper are as follows:
\begin{itemize}
    \item Introducing the first open-access dataset for agricultural potentials of three crop types.
    
    \item Providing a large collection (158 GB) of preprocessed high-resolution, multi-temporal, multi-spectral satellite imagery.
    
    \item Releasing a multi-purpose dataset that is designed for diverse machine learning and model-based tasks.

    \item Establishing baseline benchmarks to guide further development and perspectives.

\end{itemize}

\section{Limitations of Current Approaches}

\begin{table}[t]
    \centering
    \caption{Comparison of existing multispectral remote sensing datasets (Res. stands for spatial resolution).}
    \label{tab:datasets}
    \begin{tabular}{l|c|c|c|c|c}
         \multirow{2}{*}{Dataset} & Size & Res. & \multirow{2}{*}{Bands} & Temporal & Label \\
          & (patches) & (m/px) & & length & type \\
         \hline
         fMoW-S2 & 882,779 & 10 & 13 & 3 & Single  \\
         EuroSAT & 27,000 & 10 & 13 & 1 & Single \\
         PASTIS & 2,433 & 10 & 10 &  38-61 & Single  \\
         Sen4AgriNet & 225,000 & 10/20/60 & 13 & 150-250 & Single \\
         BigEarthNet & 590,326 & 10/20/60 & 12 & 1 & Multi-label\\
         \hline
          \textbf{AgriPotential} & \multirow{2}{*}{8,890} & \multirow{2}{*}{5} & \multirow{2}{*}{10} & \multirow{2}{*}{11} & Ordinal/ \\
          \textbf{(ours)}& & & & & multi-label\\
         \hline
    \end{tabular}
\end{table}

\subsection{Agricultural Potential Assessment}

In recent years, many studies have proposed machine learning frameworks to determine agricultural potentials based on environmental variables and soil characteristics \cite{ed2023predictive, el2022assessment, alnaimy2022spatio}.

These approaches typically rely on in-situ measurements such as soil pH, nutrient content, salinity, and moisture retention, which are analyzed in laboratories and then processed with traditional machine learning algorithms. For instance, \cite{ed2023predictive} made a crop type recommendation system using classification models, while \cite{el2022assessment} and \cite{alnaimy2022spatio} applied multivariate statistical methods to assess soil suitability. 

While these approaches demonstrate promising results, they are constrained by their heavy reliance on expert knowledge and intensive laboratory analysis, making them time-consuming and difficult to scale. Additionally, we did not identify any publicly available datasets that offered labeled data on agricultural potential prediction.

To extend prior work, we introduce an open-access, field-validated, remote sensing dataset that uses expert-validated labels. Our goal is to provide a scalable solution for assessing agricultural potentials without relying on repeated laboratory measurements.

\subsection{Remote Sensing Datasets} 

Despite the lack of public datasets designed for agricultural potentials, several remote sensing datasets address similar applications. By reviewing these datasets below we aim to highlight the limitations in current resources that motivated us to build AgriPotential. Similar applications include:

\textbf{Land Use Land Cover (LULC)}. Many datasets have been created for LULC applications. For instance, Functional Map of the World - Sentinel-2 (fMoW-S2)~\cite{cong_functional_2022} provides over 880,000 image patches for land use classification, offering a strong geographical and temporal diversity. BigEarthNet~\cite{sumbul2019bigearthnet} contains over 590,000 Sentinel-2 image patches across 10 European countries, annotated with multi-label land cover classes. EuroSAT~\cite{helber2019eurosat, helber2018introducing} includes 27,000 annotated Sentinel-2 patches labeled across 10 LULC classes. A simplified RGB version is also provided.

\textbf{Crop Classification}. Datasets such as PASTIS~\cite{garnot2021panoptic} and PASTIS-R~\cite{garnot2021mmfusion} are designed for crop type segmentation and panoptic mapping. PASTIS includes annotated time series from Sentinel-2, while PASTIS-R expands PASTIS with radar data from Sentinel-1, allowing for improved modeling under cloudy conditions. Sen4AgriNet~\cite{sen4agri} also offers multi-temporal crop type data for both crop type classification and segmentation.

\textbf{Yield Prediction}. Other datasets~\cite{khaki2019crop, Khaki_2020, fudong:kdd24:crop_net} such as CropNet~\cite{fudong:kdd24:crop_net} integrated Sentinel-2 imagery with historical yield values to enable yield modeling based on regression tasks. These datasets are designed to estimate actual yields rather than land mapping.

Although many of these datasets and dozens that were featured online~\footnote{\url{https://github.com/chrieke/awesome-satellite-imagery-datasets} (accessed on May 23$^{\text{rd}}$ 2025)}~\footnote{\url{https://github.com/satellite-image-deep-learning/datasets} (accessed on May 23$^{\text{rd}}$ 2025)} support machine learning tasks such as single-label and multi-label classification, as well as regression, only a subset offers both multi-spectral and multi-temporal dimensions. In contrast, AgriPotential features pixel-wise annotations suitable for segmentation and integrates both multi-spectral and multi-temporal satellite data. It is specifically well-suited for a wide range of machine learning tasks, including segmentation, ordinal classification, multi-label prediction, and regression. Beyond supporting various tasks, the dataset also enables modeling of spatial, temporal, and spectral dimensions, making it a valuable resource for a wide range of studies in remote sensing and machine learning. As shown in Table~\ref{tab:datasets}, this unique combination of properties distinguishes AgriPotential from existing remote sensing datasets, which are largely designed for categorical LULC classification tasks.

\section{AgriPotential Dataset}

\begin{figure*}[!tp]
    \centering
    \includegraphics[width=\linewidth]{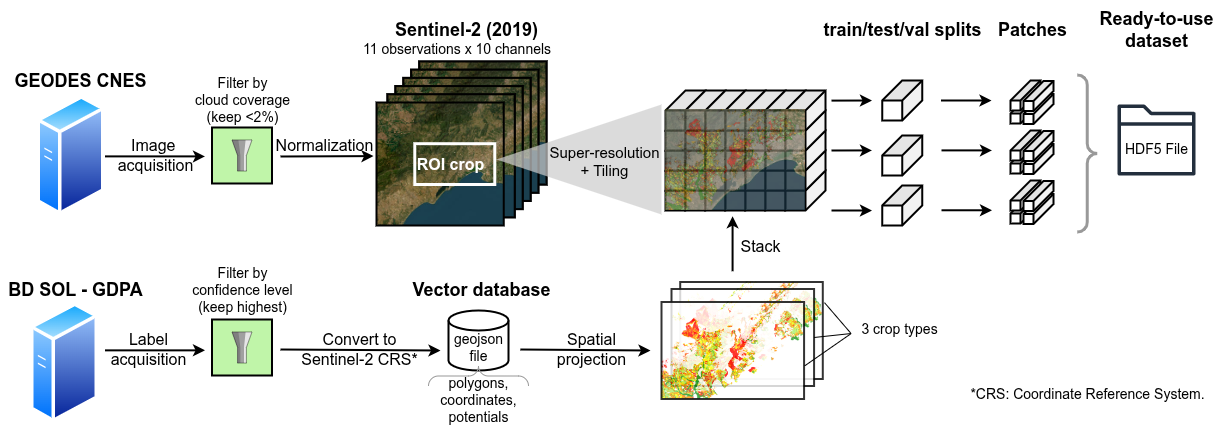}
    \caption{Overview of the AgriPotential dataset end-to-end construction pipeline. The process begins with the selection of Sentinel-2 images from the GEODES CNES Portal, filtered to retain only those with minimal cloud cover. Ground truth annotations are sourced from the BD SOL - GDPA database and filtered to include only the highest confidence levels. These annotations are provided in vector format, with polygons representing spatial regions labeled by agricultural potential. The coordinate reference system (CRS) of the ground truth is then reprojected to match that of the Sentinel-2 images to ensure an accurate spatial alignment. The two data sources are then stacked and a grid of tiles is overlaid, which are then divided into training, validation, and testing subsets. Finally, patches are extracted from the tiles to create the final dataset.}
    \label{fig:processus}
\end{figure*}

In this section, we present the AgriPotential dataset. The dataset includes high-resolution satellite images paired with field-validated agricultural potential labels. Our construction pipeline is presented in Figure~\ref{fig:processus}. 

We begin with the acquisition of Sentinel-2 images that we filter and process to retain unclouded images. Then, we acquire ground truth labels, derived from domain experts of pedoclimatic studies, that date back to 2019 and describe the potential of three crop types. We align these labels with the satellite images. We then split the data into subsets that are ready to use for machine learning applications. The following subsections describe in more details each step of this process.

\subsection{Area of study}

\begin{figure}[!htp]
    \centering
    \vspace{-5pt}
    \includegraphics[width=0.45\linewidth]{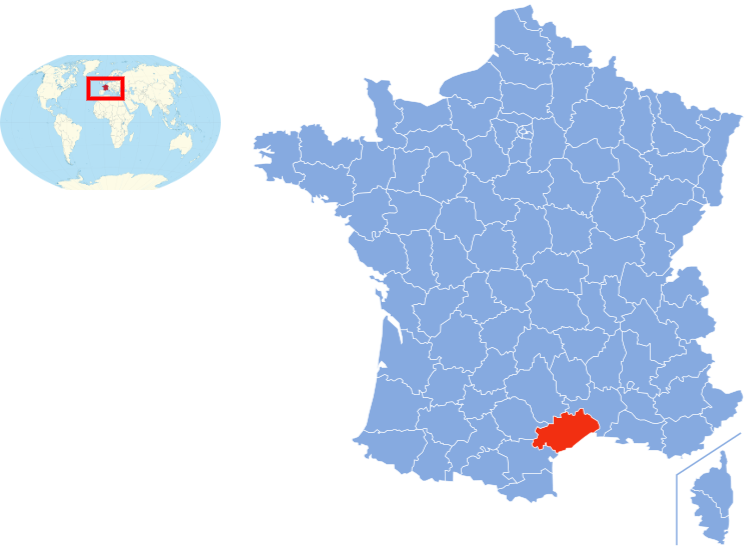}
    \caption{Hérault Department in South of France. 
    }
    \label{fig:herault}
\end{figure}

The AgriPotential Dataset covers parts of the Hérault Department in the Occitanie Region in France (see Figure~\ref{fig:herault}). This region benefits from a Mediterranean climate, characterized by mild, wet winters, and hot, dry summers. In addition to its diverse topography, ranging from coastal plains to inland hills, the department has a rich agricultural variability.

 Agriculture is a key factor in Hérault's economy, with nearly 30\% of its surface area (more than 185,000 hectares) dedicated for crops. While most of it is cultivated with viticulture (grapevines), the rest of agricultural land is devoted to market gardening (fruits and vegetables), and field crops (e.g. cereals).

 \subsection{Agricultural Potential Data}

 To develop our dataset, we use "BD Sol - GDPA", a geospatial ground truth database made publicly~\footnote{\url{https://data.laregion.fr/explore/dataset/bd-sol-gdpa-herault\%40data-herault-occitanie} (accessed on May 28$^\text{th}$ 2025)} available by the Hérault Department. The data are licensed under the Etalab Open License v2.0, which is a French open data license that permits free, worldwide, and perpetual reuse of public information for any purpose as long as the source is attributed, and is compatible with other open licenses such as CC-BY.

\begin{figure}[!htp]
    \centering
    \includegraphics[width=0.8\linewidth]{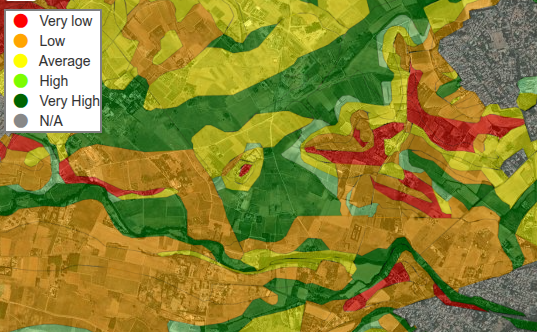}
    \caption{Agricultural potential classes ranging from "Very low" to "Very high", represented over a selected patch in the Hérault region. The colored map shows potential levels for viticulture.}
    \label{fig:potential-example}
\end{figure}

BD Sol - GDPA provides detailed agricultural potential that focus on three main agricultural sectors found in Southern France: viticulture, market gardening, and field crops. Originally created in 2013 and updated in both 2018 and 2019, the database is built on a deep analysis of pedoclimatic factors (soil depth, texture, stoniness, and water retention capacity) that classifies the potential of each crop type into five potential classes ranging from "very low" to "very high" independently of current land use. Figure~\ref{fig:potential-example} illustrates these classes using a color-based representations that visually illustrate the agricultural potentials of a subset of the full dataset. The illustrated potentials serve as the ground truth reference in our dataset.

 The data from BD Sol - GDPA are validated by domain experts and provided with confidence levels, reflecting data reliability levels. We applied a filter to keep the most reliable data to increase the robustness of the dataset.

 \subsection{Satellite Data}
 
\begin{table}[]
    \caption{Sentinel-2 multispectral bands used in the AgriPotential Dataset}
    \label{tab:sentinel2-bands}
    \centering
    \begin{tabular}{c|c|c|c}
        Band & Name & Resolution (m/px) & $\lambda$ (nm) \\
        \hline
         B2 & Blue & 10 & 490  \\
         B3 & Green & 10 & 560  \\
         B4 & Red & 10 & 665 \\
         B5 & Visible and NIR & 20 & 705  \\
         B6 & Visible and NIR & 20 & 740  \\
         B7 & Visible and NIR & 20 & 783 \\
         B8 & Visible and NIR & 10 & 842  \\
         B8A & Visible and NIR & 20 & 865 \\
         B11 & Shortwave infrared & 20 & 1610 \\
         B12 & Shortwave infrared & 20 & 2190 \\
        \hline
    \end{tabular}
\end{table}

\begin{figure}
    \centering
    \includegraphics[width=\linewidth]{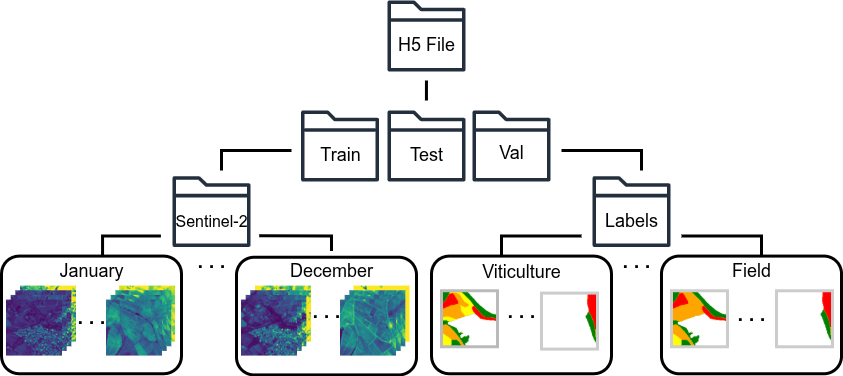}
    \caption{The hierarchical file structure of the AgriPotential Dataset.}
    \label{fig:strucutre}
\end{figure}

To collect satellite data, we use the GEODES Portal of CNES, the French National Space Agency, which provides access to a wide range of Earth Observation satellite data, including Sentinel-2. Sentinel-2 satellites are part of the European Union's Copernicus Earth Observation Program, operated by the European Space Agency (ESA), that aim to monitor land and vegetation on a global scale. 

Sentinel-2 satellites carry a MultiSpectral Instrument (MSI) that captures 13 wavelengths of light reflected from the Earth surface, ranging from visible and near-infrared to shortwave infrared, with a spatial resolution varying between 10, 20, and 60 meters per pixel depending on the spectral band. The Sentinel-2 constellation enables a revisit time of five days, allowing frequent and timely coverage of the Earth.

This data are available in various preprocessing levels, including the Level-2A (L2A) product, which provides accurate and reliable surface reflectance by applying atmospheric corrections. The GEODES CNES portal supplies 10 spectral bands (see Table~\ref{tab:sentinel2-bands}), since the remaining bands are primarily used for correction purposes and are not distributed.

\subsection{Building the Dataset}

To ensure temporal alignment with the agricultural potential labels from 2019, our initial goal was to collect one Sentinel-2 image per month throughout the year. Although multiple images are available for certain months, we deliberately limited the selection to approximately one per month to ensure a uniform and temporally balanced time series. Capturing images across the entire year, enables a more robust modeling of agricultural potentials as they depend on pedoclimatic conditions that are relatively stable over short periods of time. However, due to the persistent cloudiness in November, no images were available for that month. Sentinel-2 images are provided with a cloud mask that allows us to compute the cloud coverage. Only images with less than 2\% cloud coverage were selected to preserve spectral continuity. This monthly sampling allows us to capture seasonal variability, resulting in 11 high-quality images that span the year 2019.

All selected images are super-resolved from their native resolutions (10m/px and 20m/px) to 5m/px using a Convolutional Anchored Regression Network (CARN) that is recognized for achieving superior performance among other state-of-the-art super-resolution methods~\cite{li2018carn}. We used a CARN that was pretrained on Sen2Ven$\mu$s~\cite{michel2022sen2venmus}\footnote{\url{https://github.com/Evoland-Land-Monitoring-Evolution/sentinel2_superresolution} (accessed on August 27th 2025)}, a benchmark specifically designed for super-resolution of Sentinel-2 images. Super-resolution is common in remote sensing datasets that integrate bands of different spatial resolutions~\cite{cong_functional_2022, helber2019eurosat, garnot2021panoptic}. The motivation for this enhancement is to align all spectral bands to a common high-resolution that offers rich spatial context. Qualitative inspection confirmed that the process successfully enhanced spatial clarity without introducing obvious false patterns in images (see Supplement).

Each image is then normalized per band and per timestamp by dividing pixel values by 10,000, according to the official Sentinel-2 L2A product format specifications guide. This normalization step converts the raw reflectance values, originally ranging from 0 to 10,000 into a standardized scale between 0 and 1. Normalizing Sentinel-2 data improves numerical stability during operations and ensures that all time steps are on the same scale, making temporal comparisons more reliable and meaningful.

Before aligning the ground truth with satellite images, we filter the agricultural potential annotations to keep only those with the highest confidence levels to make our dataset more robust and reliable. Then, to align with Sentinel-2 images, we reproject the coordinates of the annotations (EPSG:4326) to the same coordinate reference system (EPSG:32631) using GDAL\footnote{Geospatial Data Abstraction Library: https://gdal.org/en/stable/ (accessed on May 27$^\text{th}$ 2025)}, a standard well-known geospatial transformation tool, in order to ensure pixel-level alignment between Sentinel-2 images and ground truth annotations. Once aligned, ground truth annotations are spatially projected to match the 5 m/px resolution of the satellite images, resulting in pixel-wise masks. Then, Sentinel-2 images and the corresponding labels are stacked together to form a unified data cube that measures 18770x15486 pixels.

The resulting data cube is divided into small tiles of 256x256 pixels that are randomly shuffled into three subsets: training, testing, and validation following 80/10/10 partitioning ratios. This tiling and shuffling step is performed before final patch extraction to prevent data leakage between different subsets. Each tile is then further divided into overlapping patches of 128x128 pixels with a 50\% stride. The overlap ensures that each pixel is provided with enough spatial context from its surroundings, which is essential in satellite imagery. During this process, we only keep the patches that contain labeled pixels since some areas in the original labels lack ground truth annotations.

\subsection{File Format}

The dataset contains multiple dimensions - spectral, temporal, spatial, and crop type - and has a large volume of 158 GB. To efficiently store, organize, and compress this complex data, we chose a hierarchical file format, HDF5 (.h5), which reduced the dataset size to 30 GB. HDF5 stores data in a hierarchical structure, using key-value pairs at each level, where the leaf nodes contain data matrices (see Figure~\ref{fig:strucutre}).

The dataset is organized by crop type, where each label is represented as a one-hot encoded vector corresponding to one of five classes of agricultural potential.
Unlabeled pixels are labeled with a special "ignore" label: (0,0,0,0,0). Each patch also includes a time series of Sentinel-2 satellite images, where each image is a monthly observation. These images are organized chronologically by month, and each image has 10 spectral bands. In the supplementary material\footnote{\url{https://zenodo.org/records/15551829}}, we provide a PyTorch example of how to load the data for different types of analysis.

\begin{figure*}[t]
    \centering
    \begin{subfigure}[b]{0.32\textwidth}
        \centering
        \includegraphics[width=\linewidth]{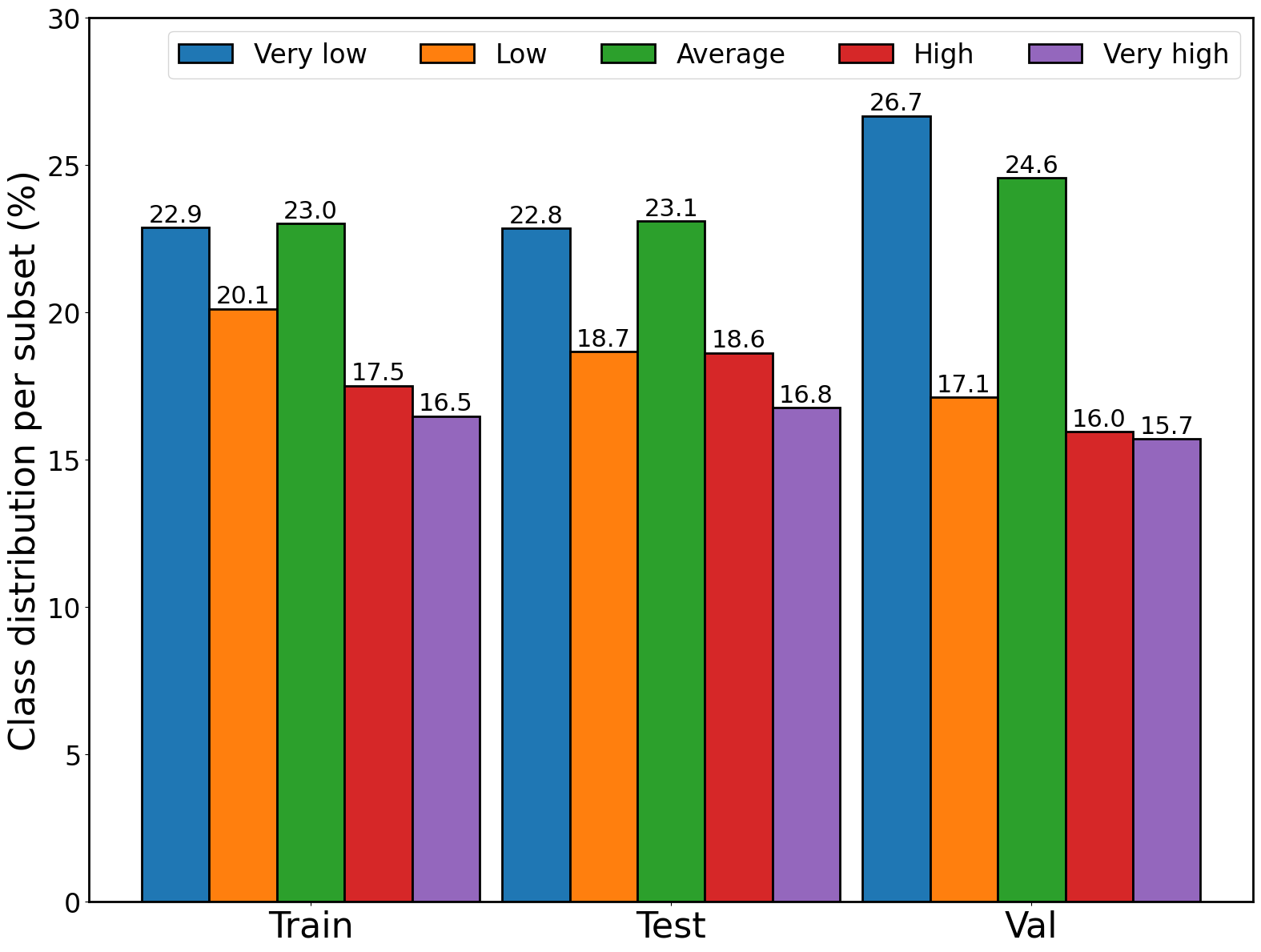}
        \caption{Viticulture}
        \label{viticulture_class}
    \end{subfigure}
    \hfill
    \begin{subfigure}[b]{0.32\textwidth}
        \centering
        \includegraphics[width=\linewidth]{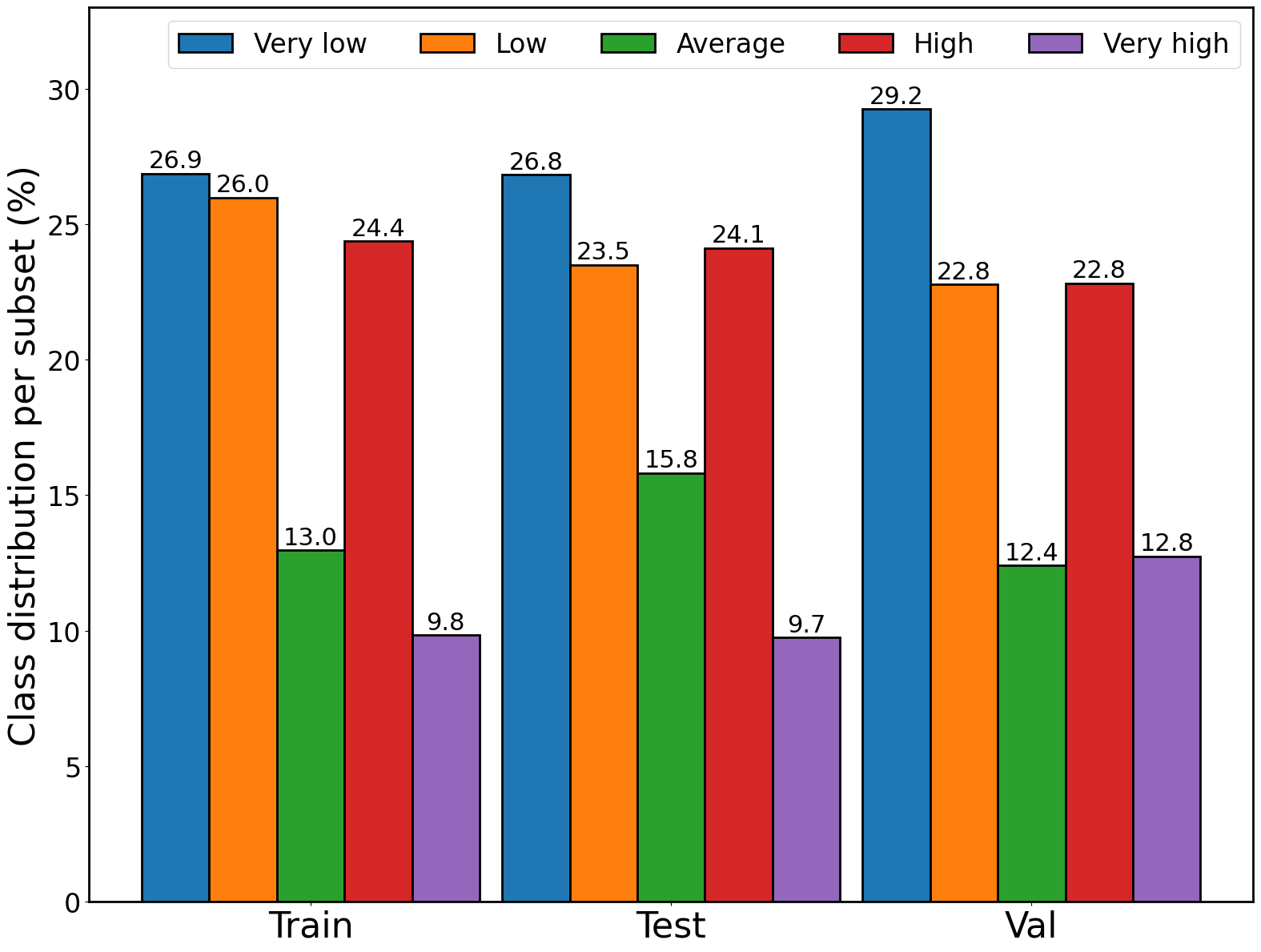}
        \caption{Market Gardening}
        \label{fig:market_class}
    \end{subfigure}
    \hfill
    \begin{subfigure}[b]{0.32\textwidth}
        \centering
        \includegraphics[width=\linewidth]{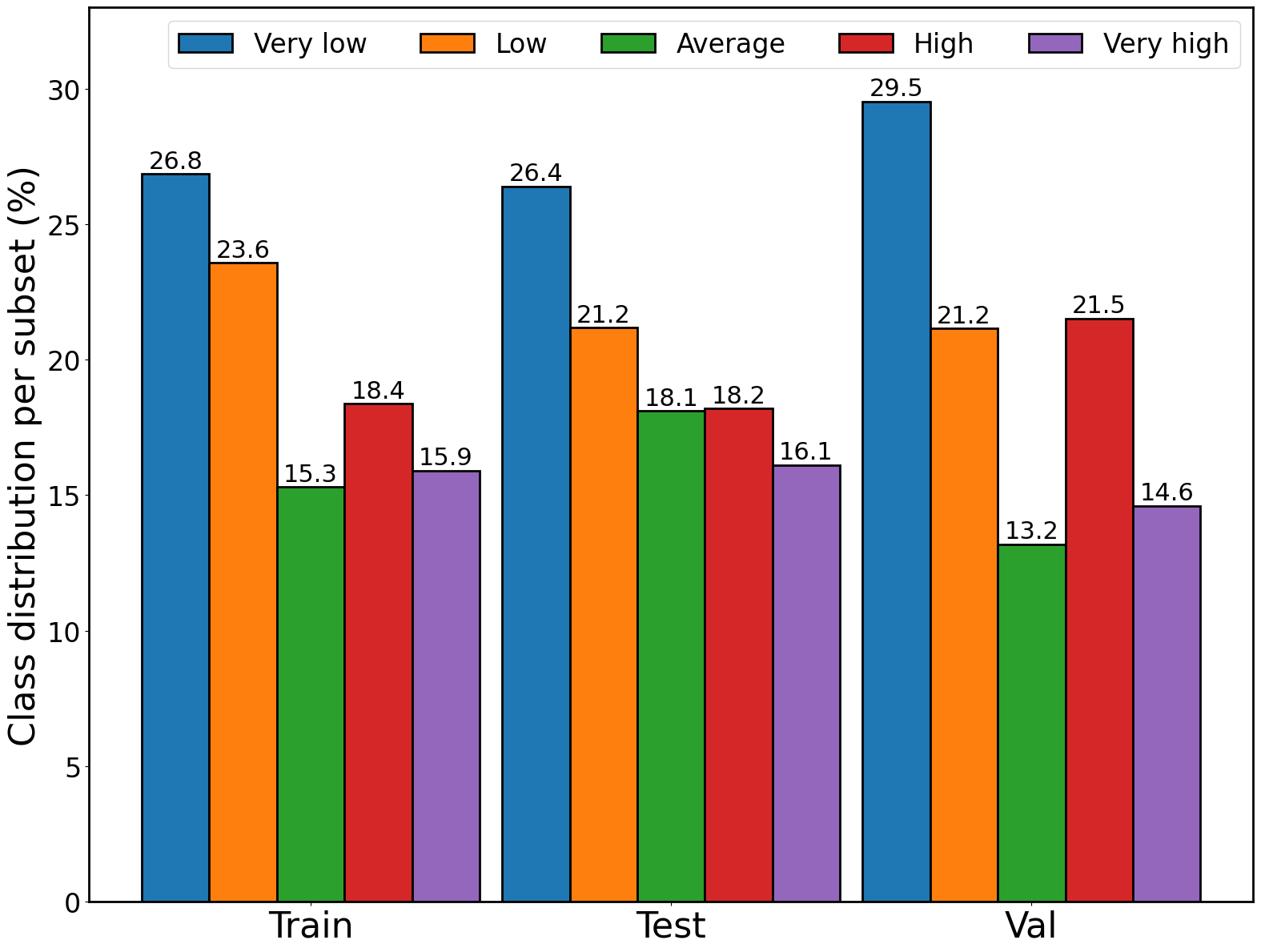}
        \caption{Field Crops}
        \label{fig:field_class}
    \end{subfigure}
    \caption{Class distribution across the three crop types. The class frequencies reveal imbalanced representations within each crop type. }
    \label{fig:distribution}
\end{figure*}

\begin{figure}[t]
    \centering
    \includegraphics[width=0.75\linewidth]{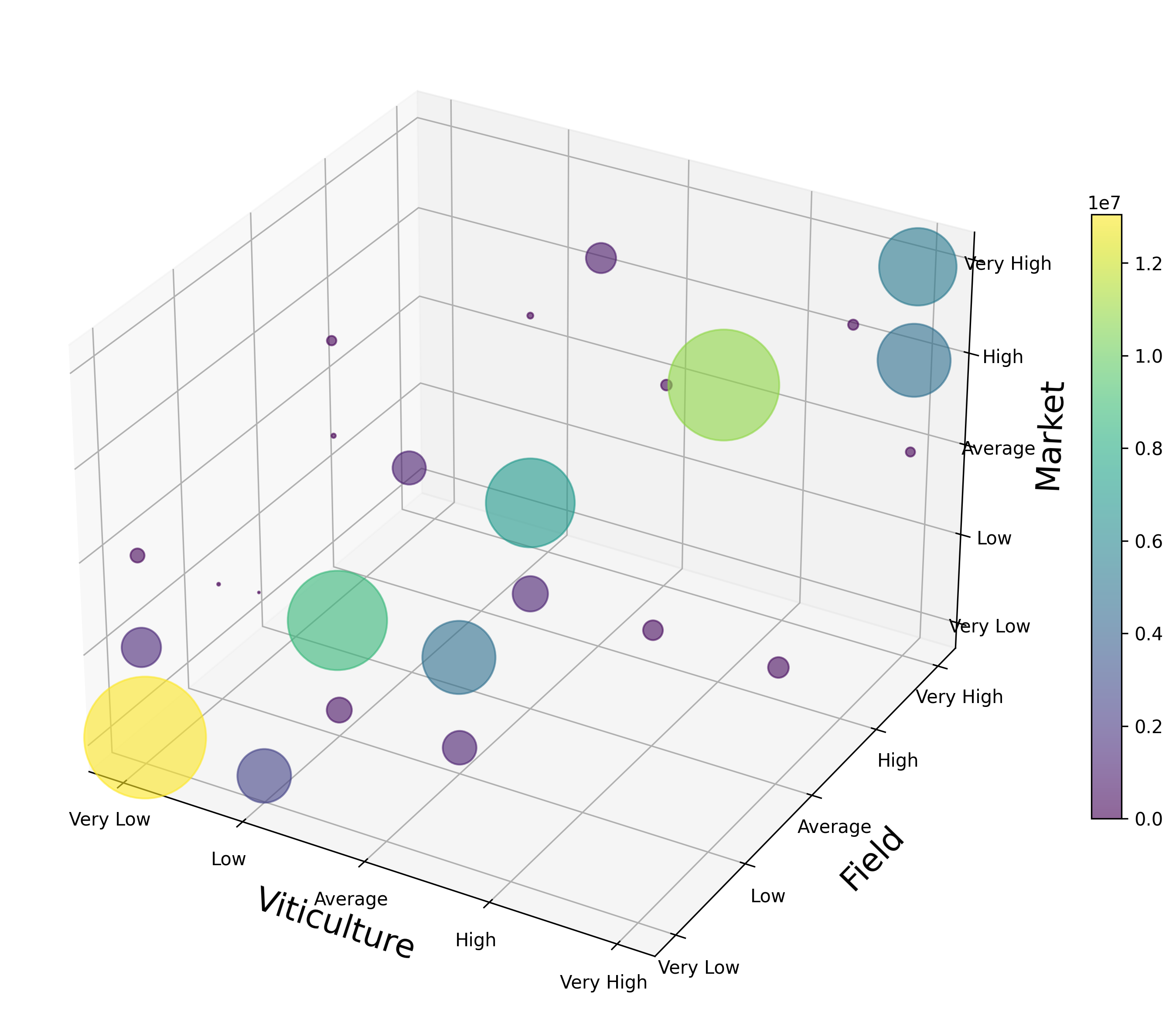}
    \caption{The frequency of co-occurrence of agricultural potential labels across the three crop types. Larger circles indicate higher co-occurrence frequencies. The same pixel can have different agricultural potentials depending on the crop type, highlighting the suitability of multi-label classification tasks.}
    \label{fig:spheres}
    \vspace{-2pt}
\end{figure}

\subsection{Data Statistics}

\subsubsection{\textbf{Data Completeness}}
The dataset contains a total of 8890 patches, split into 7095 for training, 881 for testing, and 914 for validation. However, not all pixels within these patches are labeled. This is due to the fact that the ground truth annotations, derived from the BD SOL-GDPA database, are available only for specific regions. On average, 54\% of the pixels per patch are labeled in the training set, 60\% in the validation subset, and 52\% in the test set.

\subsubsection{\textbf{Class Distribution}}

We provide the distribution of the potential levels across the three crop types or classes, which can be helpful for applications that consider class weights (see Figure~\ref{fig:distribution}). In viticulture, "average" and "high" potentials are more frequently represented, while in field crops, the dataset has higher frequencies of "low" and "very low" potentials. In market gardening, "average" and "very high" are less common. This class distribution is consistent across the training, validation, and test splits.
 
\subsubsection{\textbf{Temporal Evolution of Spectral Bands}}
To gain a deeper insight into the temporal dynamics present in the AgriPotential dataset, we analyze the temporal evolution of spectral reflectance across all 10 Sentinel-2 bands, grouped by crop type and agricultural potentials. The results, presented in the supplementary material, show a clear temporal pattern. Reflectance values generally increase from January, reach their peak between April and September, and decline toward the end of the year. Additionally, within each crop type, agricultural potential classes consistently exhibit a unique spectral signature. In general, high potential classes show higher reflectance, while lower potential classes tend to show smaller values.

\subsubsection{\textbf{Co-Occurrences of Labels across Crop Types}}

To better understand the relationships between crop-specific potentials at the pixel level, we analyze how agricultural potential labels co-occur across the three crop types. Figure~\ref{fig:spheres} illustrates the proportion of pixels where potential co-occur. A significant number of pixels have the same label for all crop types. However, in other cases, potentials can be opposite depending on the crop type (e.g. "High" for "Field" and "Very low" for "Market"). This confirms that the same area can be suitable for a crop type while being less suitable for another. As a result, each pixel can be classified into multiple labels that differ significantly, enabling the use of multi-label classification or crop recommendation systems.

\section{Experimental Validation}

To demonstrate the validity and feasibility of using our dataset for machine learning tasks, we conducted baseline experiments using a minimal approach across three representative tasks: single-class classification, regression, and ordinal regression. These tasks reflect common modeling approaches in machine learning.  We stacked all 11 Sentinel-2 observations across the 10 spectral bands into a single 110 channel input and applied a 2D Convolutional Neural Network, namely UNet~\cite{ronneberger2015u}. This setup deliberately avoids sophisticated modeling approaches to demonstrate what performance can be achieved using basic 2D convolutions alone. Full training configuration is given in the Supplementary Material.

The results demonstrated that a 2D convolutional model is capable of capturing meaningful spatial and spectral patterns across crop types. Specifically, using ordinal labels consistently outperformed other representations (one-hot and scalars) in terms of both mean absolute error (MAE) and accuracy with tolerance of $\pm$ 1 class, as these metrics reflect the practical significance of predictions.

The confusion matrices (see Supplement) reveal that when the model makes mistakes, it tends to predict neighboring classes rather than entirely incorrect ones. This behavior suggests that the model captures the correct direction of agricultural potentials. 

\section{Perspective and Limitations}
The dataset is currently limited by its geographical scope that covers only parts of Southern France. While the selected region offers diverse topography and agricultural practices, it is confined to the Mediterranean climate. A further enhancement would be to incorporate data from regions with different climate zones to improve the dataset’s diversity and enable the development of more generalizable and scalable studies.

Another limitation lies in multi-spectral imagery which, despite offering rich spectral information, is susceptible to cloud cover. This results in gaps in the time series, especially in clouded seasons. Integrating radar remote sensing data, for example from Sentinel-1 satellites, would address this issue.

A further challenge to be considered is the potential confirmation bias in model evaluation. While the labels are independent of current land use, a model might still learn shortcuts by associating the presence of a specific crop with its high potential label. However, this is not universally the case, because the label co-occurrence analysis presented in Figure~\ref{fig:spheres} shows that many pixels share the same potential level across all crop types, suggesting that high potential land often exhibits  general suitability rather than being crop-specific. Nevertheless, future work could focus on guiding the model to avoid such shortcuts. Approaches could include techniques such as spectral unmixing, which decomposes pixel signals into separate components (e.g. soil, plants, minerals).

In our future perspectives, we aim to further enhance the dataset by incorporating additional remotely sensed data sources, such as weather data, and thermal satellite data. This approach would preserve the remote sensing nature of the dataset, while enabling richer environmental context.

In parallel, we aim to develop solutions for multi-spectral multi-temporal image processing, which is an emerging area of research that has been gaining significant attention in recent literature~\cite{garnot2021panoptic, hong2024spectralgpt}. Exploring the integration of transformer-based models for temporal and spectral attention also represents a promising direction.

\section{Conclusion}
In this work, we introduced AgriPotential, a remote sensing dataset based on multi-spectral and multi-temporal data. By integrating ground truth annotated by domain experts with high-resolution Sentinel-2 images, AgriPotential addresses the need for scalable and data-driven approaches to agricultural potential prediction. The dataset supports a wide range of machine learning tasks and serves as a robust benchmark for future research in smart agriculture and remote sensing.

\section*{Acknowledgments}

We thank the Department of Hérault for publicly releasing the BD Sol - GDPA dataset. Their open data initiative and transparent access to high-quality agricultural information greatly facilitated the completion of this work. The work benefits from European Union funds through the AI4AGRI project entitled “Romanian . Excellence Center on Artificial Intelligence on Earth Observation Data for Agriculture” that received 
funding from the European Union’s Horizon Europe research and innovation program under the 
grant agreement no. 101079136. 
\begin{figure}[h]
    \centering
    \includegraphics[width=\linewidth]{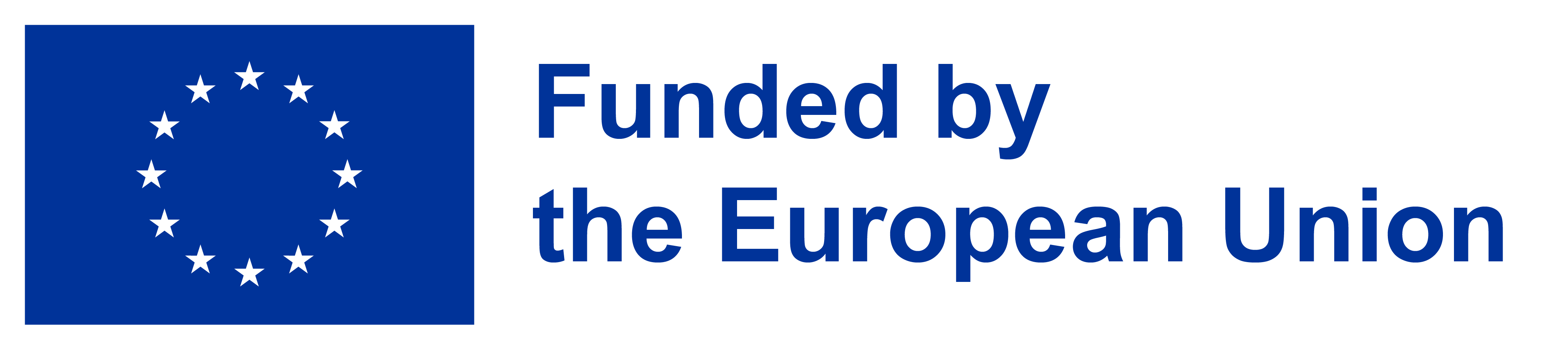}
\end{figure}




\bibliographystyle{IEEEtran}
\bibliography{references}

\end{document}